\documentclass[letterpaper]{article}
\usepackage{aaai19}
\usepackage{times}
\usepackage{helvet}
\usepackage{courier}
\usepackage{url}
\usepackage{graphicx}
\usepackage{amssymb}
\usepackage{color}
\usepackage{comment}

\usepackage{booktabs}
\usepackage{amsfonts}
\usepackage{amsmath}
\usepackage{amssymb}
\usepackage{makecell}
\usepackage{multirow}
\usepackage{pbox}
\usepackage{mathtools}
\usepackage[ruled, vlined, linesnumbered, noend]{algorithm2e}
\usepackage{tikz}
\usepackage{subfigure}
\usepackage{comment}
\usepackage{enumitem}
\usepackage{arydshln}

\usepackage{multirow}

\usepackage{afterpage}

\usepackage{graphicx,xspace}
\usepackage{epstopdf}
\usepackage{amsthm}

\theoremstyle{definition}
\newtheorem{definition}{Definition}

\theoremstyle{definition}
\newtheorem{exmp}{Example}[section]

\title{Embedding Uncertain Knowledge Graphs}
\author{Xuelu Chen, Muhao Chen, Weijia Shi, Yizhou Sun, Carlo Zaniolo\\
University of California, Los Angeles\\
Los Angeles, CA, 90095, USA\\
\{shirleychen, muhaochen, swj0419, yzsun, zaniolo\}@cs.ucla.edu}

\begin{document}

\maketitle

\begin{abstract}
Embedding models for deterministic Knowledge Graphs (KG) have been extensively studied, with the purpose of capturing latent semantic relations between entities and incorporating the structured knowledge they contain into machine learning.
However, there are many KGs that model uncertain knowledge, which typically model the inherent uncertainty of relations facts with a confidence score,
and embedding such uncertain knowledge represents an unresolved challenge.
The capturing of uncertain knowledge will benefit many knowledge-driven applications such as question answering and semantic search by providing more natural characterization of the knowledge.
In this paper, we propose a novel uncertain KG embedding model \texttt{UKGE}, 
which aims to preserve both structural and uncertainty information of relation facts in the embedding space. 
Unlike previous models that characterize relation facts with binary classification techniques, 
\texttt{UKGE}~learns embeddings according to the confidence scores of uncertain relation facts.
To further enhance the precision of \texttt{UKGE}, 
we also introduce probabilistic soft logic to infer confidence scores for unseen relation facts during training.
We propose and evaluate two variants of \texttt{UKGE}~based on different confidence score modeling strategies.
Experiments are conducted on three real-world uncertain KGs via three tasks, i.e. confidence prediction, relation fact ranking, and relation fact classification.
\texttt{UKGE}~shows effectiveness in capturing uncertain knowledge by achieving promising results, and it consistently outperforms baselines on these tasks.
\end{abstract}


\section{Introduction}
Knowledge Graphs (KGs) provide structured representations of real-world entities and relations, which are categorized into the following two types: (i) \emph{Deterministic KGs}, such as YAGO \cite{rebele2016yago} and FreeBase \cite{bollacker2008freebase}, consist of deterministic relation facts that describe semantic relations between entities; (ii) \emph{Uncertain KGs} including ProBase \cite{wu2012probase}, ConceptNet \cite{speer2017conceptnet} and NELL \cite{mitchell2018never}
associate every relation fact with a confidence score that represents the likelihood of the relation fact to be true. 

KG embedding models are essential tools for incorporating the structured knowledge representations in KGs into machine learning.
These models encode entities as low-dimensional vectors and relations as algebraic operations among entity vectors. They accurately capture the similarity of entities and preserve the structure of KGs in the embedding space.
Hence, they have been the crucial feature models that benefit numerous knowledge-driven tasks \cite{bordes2014open,he2017symmetric,das2018go}. Recently, extensive efforts have been devoted into embedding deterministic KGs. Translational models, e.g., TransE \cite{bordes2013translating} and TransH \cite{wang2014knowledge}), and bilinear models, e.g. DistMult \cite{yang2014embedding} and ComplEx \cite{trouillon2016complex}, have achieved promising performance in many tasks, such as link prediction \cite{yang2014embedding,trouillon2016complex},  relation extraction \cite{westonconnecting}, relational learning \cite{nickel2016holographic},
and ontology population \cite{chen2018on2vec}.

While current embedding models focus on capturing deterministic knowledge, 
it is critical to incorporate uncertainty information into knowledge sources for several reasons.
First, uncertainty is the nature of many forms of knowledge.
An example of naturally uncertain knowledge is the interactions between proteins. 
Since molecular reactions are random processes, biologists label the protein interactions with their probabilities of occurrence and present them as uncertain KGs called Protein-Protein Interaction (PPI) Networks. 
Second, uncertainty enhances inference in knowledge-driven applications. For example, short text understanding often entails interpreting real-world concepts that are ambiguous or intrinsically vague. The probabilistic KG Probase \cite{wu2012probase} provides a prior probability distribution of concepts behind a term that has critically supported short text understanding tasks involving disambiguation \cite{wang2015inference,wang2016understanding}. 
Furthermore, uncertain knowledge representations have largely benefited various applications, such as question answering \cite{yih2013question} and named entity recognition \cite{ratinov2009design}.

Capturing the uncertainty information 
with KG embeddings remains an unresolved problem.
This is a non-trivial task for several reasons.
First, compared to deterministic KG embeddings,
uncertain KG embeddings need to encode additional confidence information to preserve uncertainty.
Second, current KG embedding models cannot capture the subtle uncertainty of unseen relation facts, as they assume that all the unseen relation facts are false beliefs and minimize the plausibility measures of relation facts. One major challenge of learning embeddings for uncertain KGs is to properly estimate the uncertainty of unseen relation facts. 

To address the above issues, we propose a new embedding model \texttt{UKGE}~(\texttt{\underline{U}ncertain \underline{K}nowledge \underline{G}raph \underline{E}mbeddings}), which aims to preserve both structural and uncertainty information of relation facts in the embedding space. Embeddings of entities and relations on uncertain KGs are learned according to confidence scores. 
Unlike previous models that characterize relation facts with binary classification techniques, 
\texttt{UKGE}~learns embeddings according to the confidence scores of uncertain relation facts.
To further enhance the precision of \texttt{UKGE}, 
we also introduce probabilistic soft logic to infer the confidence score for unseen relation facts during training.
We propose two variants of \texttt{UKGE}~based on different embedding-based confidence functions.
We conducted extensive experiments using three real-world uncertain KGs on three tasks: (i) \emph{confidence prediction}, which seeks to predict confidence scores of unseen relation facts; 
(ii) \emph{relation fact ranking}, which focuses on retrieving tail entities for the query $(h, r, \underline{?t})$ and ranking these retrieved tails in the right order; and
(iii) \emph{relation fact classification}, which decides whether or not a given relation fact is a \emph{strong} relation fact. 
Our models consistently outperform the baseline models in these experiments.

The rest of the paper is organized as follows. We first review the related work in Section 2, then provide the problem definition and propose our model \texttt{UKGE} in the two sections that follow. In section 5, we present our experiments. Then we conclude the paper in Section 6.


\section{Related Work}
To the best of our knowledge, there has been no previous work on learning embeddings for uncertain KGs. We hereby discuss the following three lines of work 
that are closely related to this topic.

\subsubsection{Deterministic Knowledge Graph Embeddings}
Deterministic KG embeddings have been extensively explored by recent work. These models encode entities as low-dimensional vectors and relations as algebraic operations among entity vectors. 
There are two representative families of models, i.e. translational models and bilinear models.

Translational models share a common principle $\boldsymbol{h}_r+\boldsymbol{r} \approx \boldsymbol{t}_r$, where $\boldsymbol{h}_r$, $\boldsymbol{t}_r$ are the entity embeddings projected in a relation-specific space. 
The forerunner of this family, TransE \cite{bordes2013translating}, lays $\boldsymbol{h}_r$ and $\boldsymbol{t}_r$ in a common space as $\boldsymbol{h}$ and $\boldsymbol{t}$ with regard to any relation $r$. Variants of TransE, such as TransH \cite{wang2014knowledge}, TransR \cite{lin2015learning}, TransD \cite{ji2015knowledge}, and TransA, \cite{jia2016locally} differentiate the translations of entity embeddings in different language-specific embedded spaces based on different forms of relation-specific projections. Despite its simplicity, translational models achieve promising performance on knowledge completion and relation extraction tasks.

Bilinear models \cite{jenatton2012latent} model relations as the second-order correlations between entities, using the scoring function $f(h, r, t) = \boldsymbol{h}^T \boldsymbol{W_r} \boldsymbol{t}$. This function is first adopted by RESCAL \cite{nickel2011three}, a collective matrix factorization model. DistMult \cite{yang2014embedding} constrains $\boldsymbol{W_r}$ as a diagonal matrix which reduces the computing cost and also enhances the performance.
ComplEx adjusts the corresponding scoring mechanism to a complex conjugation in a complex embedding space.

There are also other models for deterministic KG embedding, 
such as neural models like Neural Tensor Network (NTN) \cite{socher2013reasoning} and ConvE \cite{dettmers2017convolutional}, and the circular-correlation-based model HolE \cite{nickel2016holographic}.

\subsubsection{Uncertain Knowledge Graphs}
Uncertain KGs provide a confidence score along with every relation fact. The development of relation extraction and crowdsourcing in recent years enabled the construction of large-scale uncertain knowledge bases. 
ConceptNet \cite{speer2017conceptnet} is a multilingual uncertain KG for commonsense knowledge that is collected via crowdsourcing. The confidence scores in ConceptNet mainly come from the co-occurrence frequency of the labels in crowdsourced task results. 
Probase \cite{wu2012probase} is a universal probabilistic taxonomy built by relation extraction.
Every fact in Probase is associated with a joint probability $P_{isA}(x, y)$.
NELL \cite{mitchell2018never} collects relation facts by reading web pages and learns their confidence scores from semi-supervised learning with the Expectation-Maximum (EM) algorithm.
Aforementioned uncertain KGs have enabled numerous knowledge-driven applications.
For example, \citeauthor{wang2016understanding} \shortcite{wang2016understanding} utilize Probase to help understand short texts. 

One recent work has proposed a matrix-factorization-based approach to embed uncertain networks \cite{hu2017embedding}. However, it cannot be generalized to embed uncertain KGs because this model only considers the node proximity in the networks with no explicit relations and only generates node embeddings. 
As far as we know, we are among the first to study the uncertain KG embedding problem.

\subsubsection{Probabilistic Soft Logic}
Probabilistic soft logic (PSL) \cite{kimmig2012short} is a framework for probabilistic reasoning. A PSL program consists of a set of first-order logic rules with conjunctive bodies and single literal heads. 
PSL takes the confidence from interval $[0,1]$ as the \textit{soft truth values} for every atom. It uses \textit{Lukasiewics t-norm} \cite{lukasiewicz2008managing} to determine to which degree a rule is satisfied. 
In combination with Hinge-Loss Markov Random Field (HL-MRF), PSL is  widely used in probabilistic reasoning tasks, such as social-trust prediction and preference prediction \cite{bach2013hinge,bach2017hinge}.
In this paper, we adopt PSL to enhance the embedding model performance on the unseen relation facts.

\section{Problem Definition}
We define the uncertain KG embedding problem in this section by first providing the definition of uncertain KGs. 

\begin{definition}{\textbf{Uncertain Knowledge Graph.}}
An uncertain KG represents knowledge as a set of relations ($\mathcal{R}$) defined over a set of entities ($\mathcal{E}$). It consists of a set of weighted triples $\mathcal{G}=\{(l,s_l)\}$. For each pair $(l, s_l)$, $l=(h, r, t)$ is a triple representing a relation fact where $h, t \in \mathcal{E}$ (the set of entities) and $r \in \mathcal{R}$ (the set of relations), and $s_l \in [0,1]$ represents the confidence score for this relation fact to be true. 
\end{definition} 

Note that we assume the confidence score $s_l \in [0,1]$ and interpret it as a probability to leverage probabilistic soft logic-based inference. The range of original confidence scores for some uncertain KG (e.g., ConceptNet) may not fall in $[0,1]$, and normalization will be needed in these cases.
Some examples of weighted triples 
are listed below. 
\begin{exmp} \label{psl_uncertain_rl} \textbf{Weighted triples.}
\begin{enumerate}
\item (choir, \texttt{relatedto}, sing): 1.00
\item (college, \texttt{synonym}, university): 0.99
\item (university, \texttt{synonym}, institute): 0.86
\item (fork, \texttt{atlocation}, kitchen): 0.4
\end{enumerate}
\end{exmp}

\begin{definition} \textbf{Uncertain Knowledge Graph Embedding Problem.} Given an uncertain KG $\mathcal{G}$, the embedding model aims to encode each entity and relation in a low-dimensional space in which structure information and confidence scores of relation facts are preserved.
\end{definition}
Notation wise, boldfaced $\boldsymbol{h}, \boldsymbol{r}, \boldsymbol{t}$ are used to represent the embedding vectors for head $h$, relation $r$ and tail $t$ respectively. 
$\boldsymbol{h}, \boldsymbol{r}, \boldsymbol{t}$ are assumed lie in $\mathbb{R}^k$.

\section{Modeling}
In this section, we propose our model for uncertain KG embeddings. 
The proposed model \texttt{UKGE}~ encodes the KG structure according to the confidence scores for both \emph{observed} and \emph {unseen} relation facts, such that the embeddings of relation facts with higher confidence scores receive higher plausibility values.

We first design relation fact confidence score modeling based on embeddings of entities and relations, 
then introduce how probabilistic soft logic can be used to infer confidence scores for unseen relations, and lastly describe the joint model \texttt{UKGE}~and its two variants.

\subsection{Embedding-based Confidence Score Modeling for Relation Facts}\label{sec:plm}
Unlike deterministic KG embedding models, uncertain KG embedding models need to explicitly model the confidence score for each triple and compare the prediction with the true score. We hereby first define and model the plausibility of triples, which can be considered as a unnormalized confidence score.

\begin{definition} \textbf{Plausibility.}
Given a relation fact triple $l$, the plausibility $g(l) \in R$ measures how likely this relation fact holds. 
The higher plausibility value corresponds to the higher confidence score $s$.
\end{definition}
Given a triple $l=(h,r,t)$ and their embeddings $\boldsymbol{h}, \boldsymbol{r}, \boldsymbol{t}$, we model the plausibility of $(h,r,t)$ by the following function:
{
\small
\begin{equation}
g(l) = \boldsymbol{r} \cdot (\boldsymbol{h} \circ \boldsymbol{t})
\end{equation}
}where $\circ$ is the element-wise product, and $\cdot$ is the inner product.
This function captures the relatedness between embeddings $\boldsymbol{h}$ and $\boldsymbol{t}$ under the condition of relation $r$ and is first adopted by DistMult \cite{yang2014embedding}.
We employ this triple modeling technique for three reasons:
(i) This technique has represented the state-of-the-art performance for modeling deterministic KGs \cite{kadlec2017knowledge},
(ii) It agrees with the nature of our model to quantify the confidence of an uncertain relation fact by comparing the relation embeddings with the pair of head and tail embeddings,
(iii) It does not introduce additional parameter complexity to the model like other techniques, such as TransH \cite{wang2014knowledge}, TransR \cite{lin2015learning}, ConvE \cite{dettmers2017convolutional} and ProjE \cite{shi2017proje}.
Nevertheless, this scoring function can be further explored in future work.

\subsubsection{From plausibility to confidence scores} 
In order to transform plausibility scores to confidence scores, we consider two different mapping functions and test them in the experimental section.
Formally, let a triple be $l$ and its plausibility score be $g(l)$, a transformation function $\phi(\cdot)$ maps $g(l)$ to a confidence score $f(l)$.
\small
\begin{equation}
\label{phi}
f(l) = \phi(g(l)), \phi:  \mathbb{R} \rightarrow [0,1]
\end{equation}
\normalsize
Two choices of mapping $\phi$ are listed below.

\noindent \textbf{Logistic function.}
One way to map plausibility values to confidence score is a logistic function as follows:
{\small
\begin{equation} \label{logistic}
\phi(x) = \frac{1}{1+e^{-(\mathbf{w} x+\mathbf{b})}}
\end{equation}
}
\noindent \textbf{Bounded rectifier.}
Another mapping is a bounded rectifier \cite{chen2015marginalizing}:

{\small
\begin{equation}
\label{linear}
\phi(x) = \min(\max(\mathbf{w} x+\mathbf{b}, 0), 1)
\end{equation}
}where $\mathbf{w}$ is a weight $\mathbf{b}$ is a bias.

\subsection{PSL-based Confidence Score Reasoning for Unseen Relation Facts} \label{sec:psl}
In order to better estimate confidence scores, both observed and unseen relation facts in KGs should be utilized. 
Deterministic KG embedding methods assume that all unseen relation facts are false beliefs, and use negative sampling to add some of these false relations into training. 
One major challenge of learning embeddings for uncertain KGs, however, is to properly estimate the uncertainty of unseen triples, as simply treating their confidence score as $0$ can no longer capture the subtle uncertainty. For example, it is common that a Protein-Protein Interaction Network KG may have no interaction records for two proteins that can be potentially binded. Ignoring such possibility will result in information loss.  

We thus introduce probabilistic soft logic (PSL) \cite{kimmig2012short} to infer confidence scores for these unseen relation facts to further enhance the embedding performance. PSL is a framework for confidence reasoning that propagates confidence of existing knowledge to unseen triples using soft logic.

\subsubsection{Probabilistic Soft Logic}
A PSL program consists of a set of first order logic rules that describe logical dependencies between facts (atoms).
One example of logical rule is shown below:
\begin{exmp} \label{rule_ex}  \textbf{A Logical Rule on Transitivity of Synonym Relation.}\\
(\underline{A},\texttt{synonym},\underline{B}) $\wedge$ (\underline{B},\texttt{synonym},\underline{C}) $\rightarrow$
(\underline{A},\texttt{synonym},\underline{C})
\end{exmp}
This logical rule describes the transitivity of the relation \texttt{synonym}. In this logical rule, \underline{A}, \underline{B} and \underline{C} are placeholders for entities,  
\texttt{synonym} is the predicate that corresponds to the relation in uncertain KGs, 
(\underline{A},\texttt{synonym},\underline{B}) $\wedge$ (\underline{B},\texttt{synonym},\underline{C}) is the body of the rule, and (\underline{A},\texttt{synonym},\underline{C}) is the head of the rule.

A logical rule serves as a template rule. By replacing the placeholders in a logical rule with concrete entities and relations, we can get rule instances, which are called \emph{ground rules}. 
Considering Example \ref{rule_ex} and uncertain relation facts from Example \ref{psl_uncertain_rl}, we can have the following ground rule by replacing the placeholders with real relation facts in KG.
\begin{exmp} \label{Ex:groundrule} \textbf{A Ground Rule on Transitivity of Synonym.}\\
({college}, \texttt{synonym}, {university}) $\wedge$ ({university}, \texttt{synonym}, {institute})
$\rightarrow$ ({college}, \texttt{synonym}, {institute})
\end{exmp}

Different from Boolean logic, PSL associates every atom, i.e., a triple $l$, with a \emph{soft truth value} from the interval $[0,1]$, 
which corresponds to the confidence score in our context and enables fuzzy reasoning. The assignment process of soft truth values is called an \emph{interpretation}. We denote the soft truth value of an atom $l$ assigned by the interpretation $I$ as $I(l)$. Naturally, for observed relation facts, their observed confidence scores are used for assignment; and for unseen triples, the embedding-based estimated confidence scores will be assigned to them:
{\small
\begin{equation} \label{I_l}
\begin{aligned}
I(l) &= s_l,  l \in \mathcal{L}^{+} \\
I(l) &= f(l), l \in \mathcal{L}^{-}
\end{aligned}
\end{equation}
}where $\mathcal{L}^{+}$ denotes the observed triple set, $\mathcal{L}^{-}$ denotes the unseen triples, $s_l$ denotes the confidence score for observed triple $l$, and $f(l)$ denotes the embedding-based confidence score function for $l$. 

In PSL, Lukasiewicz t-norm is used to define the basic logical operations, including logical conjunction ($\wedge$), disjunction ($\vee$), and negation ($\neg$), as follows:
{\small
\begin{align} 
l_1 \wedge l_2 &= \max\{0, I(l_1)+I(l_2)-1\} \label{Luk_conj} \\
l_1 \vee l_2 &= \min\{1, I(l_1)+I(l_2)\}  \label{Luk_disc} \\
\neg l_1 &= 1-I(l_1)  \label{Luk_neg}
\end{align}
}For example, according to Eq. (\ref{Luk_conj}) and (\ref{Luk_disc}), $0.8\wedge 0.3=0.1$ and $0.8 \vee 0.3= 1$. For a rule $\gamma \equiv\gamma_{body} \rightarrow \gamma_{head}$, as it can be written as $\neg \gamma_{body} \vee \gamma_{head}$, its value $p_\gamma$ can be computed as 
{\small
\begin{equation}\label{Luk_impl}
p_{\gamma_{body} \rightarrow \gamma_{head}}=\min\{1, 1-I(\gamma_{body})+I(\gamma_{head})\}
\end{equation}
}PSL regards a rule $\gamma$ as \textit{satisfied} when the truth value of its head $I(\gamma_{head})$ is the same or higher than its body $I(\gamma_{body})$, i.e., when its value is greater than or equal to 1.
{\small
\begin{equation} \label{d_r}
d_{\gamma} = 1-p_\gamma=\max\{ 0, I(\gamma_{body})-I(\gamma_{head}) \}
\end{equation}
}
Consider Example \ref{Ex:groundrule}. Let ({college}, \texttt{synonym}, {university}) be $l_1$, ({university}, \texttt{synonym}, {college})  be $l_2$, and (college, \texttt{synonym}, institute)  be $l_3$. 
Assuming that $l_1$ and $l_2$ are observed triples in KG, and $l_3$ is unseen, according to Equation (\ref{I_l}), (\ref{Luk_conj}), and (\ref{Luk_impl}), the distance to satisfaction of this ground rule is calculated as below:
{\small
\begin{equation*} \label{compute}
\begin{aligned}
d_{\gamma} & =\max\{ 0, I(l_1 \wedge l_2 )-I(l_3)\} \\
&= \max\{ 0, s_{l_1} + s_{l_2}-1-f(l_3)\} \\
& = \max\{0, 0.85-f(l_3)\}\\
\end{aligned}
\end{equation*}
}
where $s_{l_1}$ and $s_{l_2}$ are the ground truth confidence scores of corresponding relation facts in the uncertain KG.

This equation indicates that the ground rule in Example \ref{Ex:groundrule} is completely satisfied when $f(l_3)$, the estimated confidence score of (college, \texttt{synonym} institute), is above 0.85. When $f(l_3)$ is under 0.85, the smaller $f(l_3)$ is, the larger loss we have.
In other words, a bigger confidence score is preferable.
In the above example, we can see that the embedding-based confidence score for this unseen relation fact, $f(l_3)$, will affect the loss function, and it is desirable to learn embeddings that minimize these losses. Note that if we simply treat the unseen relation $l_3$ as false and use MSE (Mean Squared Error) as the loss, 
the loss would be $f(l_3)^2$, which is in favor of a lower confidence score mistakenly.  

Moreover, we add a rule to penalize the predicted confidence scores of all unseen relation facts, which can be considered as a prior knowledge, i.e., any unseen relation fact has a low probability to be true. 
Formally, for an unseen relation fact $l=(h, r, t) \in \mathcal{L}^-$, we have a ground rule $\gamma_0$:
\small
\begin{equation} \label{gamma0}
\gamma_0: \neg l
\end{equation}
\normalsize

According to Eq. (\ref{Luk_neg}) and (\ref{d_r}), its distance to satisfaction $d_{\gamma_0}$ is derived as:
\small
\begin{equation} 
d_{\gamma_0} =  f(l)
\end{equation}
\normalsize

 \subsection{Embedding Uncertain KGs} \label{sec:learning}
In this subsection, we present the objective function of  uncertain KG embeddings.

\subsubsection{Loss on observed relation facts } \label{sec_pos}
Let $\mathcal{L}^{+}$ be the set of observed relation facts,
the goal is to minimize the mean squared error (MSE) between the ground truth confidence score $s_l$ and our prediction $f(l)$ for each relation $l \in \mathcal{L}^+$:
\small
\begin{equation} \label{L+}
\mathcal{J}^+ = \sum_{l \in \mathcal{L}^{+}} |f(l)-s_l|^2
\end{equation}
\normalsize
\subsubsection{Loss on unseen relation facts}
Let $\mathcal{L}^{-}$ be the sampled set of unseen relations, and $\Gamma_l$ be the set of ground rules with $l$ as the rule head, the goal is to minimize the distance to rule satisfaction for each triple $l$. In particular, we choose to use the square of the distance as the following loss \cite{bach2013hinge}: 
{\small
\begin{equation} \label{L-}
\mathcal{J}^- = \sum_{l \in \mathcal{L}^-} \sum_{\gamma \in \Gamma_{l}} |\psi_{\gamma}(f(l))|^2
\end{equation}
}where $\psi_\gamma(f(l))$ denotes the weighted distance to satisfaction $w_\gamma d_\gamma$ of the rule $\gamma$ as a function of $f(l)$ where $w_\gamma$ is a hand-crafted weight for the rule $\gamma$.

Note that when $l$ is only covered by $\gamma_0: \neg l$, we have $\sum_{\gamma \in \Gamma_{l}} |\psi_{\gamma}(f(l))|^2=|f(l)|^2$, which is essentially the MSE loss by treating unseen relation facts as false.

\subsubsection{The Joint Objective Function}
Combining Eq. (\ref{L+}) and (\ref{L-}), we obtain the following joint objective function:
\small
\begin{equation}
\begin{aligned}
\mathcal{J} 
&= \sum_{l \in \mathcal{L}^{+}} 
|f(l)-s_l|^2
	+ \sum_{l\in\mathcal{L}^{-}}\sum_{\gamma \in \Gamma_{l}} |\psi_{\gamma}(f(l))|^2
\end{aligned}
\end{equation}
\normalsize
Similar to deterministic KG embedding algorithms, we sample unseen relations by corrupting the head and the tail for observed relation facts to generate $\mathcal{L}^{-}$ during training.

We give two model variants that differ in the choice of $f(l)$. We refer to the variant that adopts Equation (\ref{logistic}) as \texttt{UKGE$_{logi}$}~and name the one using Equation (\ref{linear}) as \texttt{UKGE$_{rect}$}.


\section{Experiments}

\begin{table}[t]
\centering
\begin{tabular}{c|ccccc}Dataset & \#Ent. & \#Rel. & \#Rel. Facts & Avg($s$) & Std($s$) \\
\hline
CN15k & 15,000 & 36 & 241,158 & 0.629 & 0.232 \\
NL27k & 27,221 & 404 & 175,412 & 0.797 & 0.242 \\
PPI5k & 4,999 & 7 & 271,666 & 0.415 & 0.213 \\
\end{tabular}
\caption{Statistics of the extracted datasets used in this paper. 
\textit{Ent.} denotes entities and \textit{Rel.} stands for relations. 
Avg($s$) and Std($s$) are the average and standard deviation of the confidence scores.}
\label{data_sub}
\end{table}

\begin{table*}[t]
\centering
\begin{tabular}{c|c|c}
Dataset & Logical Rules & Hit Ratio\\
\hline
\multirow{2}{*}{CN15k} & (\underline{A}, \texttt{relatedTo}, \underline{B})$\wedge$(\underline{B}, \texttt{relatedTo}, \underline{C})$\rightarrow$(\underline{A}, \texttt{relatedTo}, \underline{C}) & 37.0\%\\
& 
(\underline{A}, \texttt{causes}, \underline{B})$\wedge$(\underline{B}, \texttt{causes}, \underline{C})$\rightarrow$(\underline{A}, \texttt{causes}, \underline{C}) & 35.6\% \\
\hline
\multirow{1}{*}{NL27k} & 
(\underline{A}, \texttt{atheletePlaysForTeam},\underline{B})
$\wedge$
(\underline{A}, \texttt{athletePlaysSport}, \underline{C})
$\rightarrow$(\underline{B}, \texttt{teamPlaysSport}, \underline{C}) & 42.9\% \\
\hline
PPI5k & 
(\underline{A}, \texttt{binding}, \underline{B})$\wedge$(\underline{B}, \texttt{binding}, \underline{C})$\rightarrow$(\underline{A}, \texttt{binding}, \underline{C}) & 80.8\% \\
\end{tabular}
\caption{Examples of logical rules. \textit{Hit ratio} means the proportion of relation facts that have already existed in the KG
}
\label{psl_rules}
\end{table*}

In this section, we evaluate our models on three tasks: confidence prediction, relation fact ranking, and relation fact classification.

\subsection{Datasets} \label{dataset}
The evaluation is conducted on three datasets named as CN15k, NL27k, and PPI5k, which are extracted from ConceptNet, NELL, and the Protein-Protein Interaction Knowledge Base STRING \cite{szklarczyk2016string} respectively. CN15k matches the number of nodes with FB15k \cite{bordes2013translating}  - the widely used benchmark dataset for deterministic KG embeddings \cite{bordes2013translating,wang2014knowledge,yang2014embedding}, while NL27k is a larger dataset. PPI5k is a denser graph with fewer entities but more relation facts than the other two.
Table \ref{data_sub} gives the statistics of the datasets, and more details are introduced below.

{
\begin{table}[t]
\centering
\setlength\tabcolsep{2pt}
\begin{tabular}{c|cc|cc|cc}
Dataset & \multicolumn{2}{c}{CN15k} & \multicolumn{2}{|c}{NL27k} & \multicolumn{2}{|c}{PPI5k} \\
\hline
Metrics & MSE & MAE & MSE & MAE & MSE & MAE \\
\hline
URGE &10.32  &  22.72  & 7.48  & 11.35  & 1.44  & 6.00  \\
\hline
\texttt{UKGE}$_{n-}$ & 23.96  & 30.38   & 24.86  & 36.67 & 7.46 & 19.32\\
\texttt{UKGE}$_{p-}$ & 9.02  & 20.05  & 2.67  &7.03  & 0.96 & 4.09\\
\hline
\texttt{UKGE$_{rect}$} & \textbf{8.61} & \textbf{19.90} & \textbf{2.36} &  \textbf{6.90} & \textbf{0.95}  & \textbf{3.79}\\
\texttt{UKGE$_{logi}$} & 9.86 & 20.74 & 3.43 & 7.93 & 0.96 & 4.07\\
\end{tabular}
\caption{Mean squared error (MSE) and mean absolute error (MAE) of relation fact confidence prediction ($\times 10^{-2}$).}
\label{mse}
\end{table}
}

{
\begin{table}[!ht]
\centering
\setlength\tabcolsep{2pt}
\begin{tabular}{c|cc|cc|cc}
metrics & \multicolumn{2}{c}{CN15K} & \multicolumn{2}{|c}{NL27k} & \multicolumn{2}{|c}{PPI5k}\\
\hline
Dataset & linear & exp. & linear & exp. &linear&exp.  \\
\hline
TransE & 0.601 & 0.591  & 0.730 & 0.722 & 0.710 & 0.700 \\
DistMult & 0.689 & 0.677  & 0.911 & 0.897 & 0.894 & 0.880 \\
ComplEx & 0.723 & 0.712  & 0.921 & 0.913 & 0.896 & 0.881 \\
URGE & 0.572 & 0.570 &0.593 & 0.593 & 0.726 & 0.723\\
\hline
\texttt{UKGE}$_{n-}$ & 0.236  & 0.232   & 0.245  & 0.245 & 0.514 & 0.517\\
\texttt{UKGE}$_{p-}$ & 0.769  & 0.768  & 0.933  & 0.929  & 0.940  & 0.944 \\
\hline
\texttt{UKGE$_{rect}$} & 0.773 & 0.775 & 0.939 & 0.942 & 0.946 & 0.946\\
\texttt{UKGE$_{logi}$} & \textbf{0.789} & \textbf{0.788} & \textbf{0.955} & \textbf{0.956} & \textbf{0.970} & \textbf{0.969} \\
\end{tabular}
\caption{Mean normalized DCG for global ranking task. Here \emph{linear} stands for linear gain, and \emph{exp.} stands for exponential gain.}
\label{rank2}
\end{table}
}

{
\begin{table*}[t]
\centering
\setlength\tabcolsep{2pt}
\begin{tabular}{c|cc|cc|ccc}
Dataset & head & relation & true tail & confidence & predicted tail & predicted confidence & true confidence\\
\hline
\multirow{8}{*}{CN15k} & \multirow{4}{*}{rush} & \multirow{4}{*}{relatedto} & fast & 0.968 & fast & 0.703 & 0.968\\
 &  &  & motion & 0.709 & move & 0.623 & 0.557\\
 & & & rapid & 0.709 & hour & 0.603 & 0.654 \\
  & & & urgency & 0.709 & time & 0.601 & 0.105 \\

\cline{2-8}
 
 & \multirow{4}{*}{hotel} & \multirow{4}{*}{usedfor} & sleeping & 1.0 & relaxing & 0.858 & N/A \\
  & & & rest & 0.984 & sleeping & 0.849 & 1.0 \\
 & & & bed away from home & 0.709 & rest & 0.827 & 0.984 \\
  & & & stay overnight & 0.709 & hotel room & 0.797 & N/A \\
 
 \hline
 
 \multirow{4}{*}{NL27k} & \multirow{4}{*}{Toyota} & \multirow{4}{*}{competeswith} & Honda & 1.0 & Honda & 0.942 & 1.0 \\
 &  &  & Ford & 1.0 & Hyundai & 0.910 & 0.719 \\
  &  &  & BMW & 0.964 & Chrysler & 0.908 & N/A\\
   &  &  & General Motors & 0.930 & Nissan & 0.896 & 0.859 \\
\end{tabular}
\caption{Examples of relation fact ranking (global) results using \texttt{UKGE$_{logi}$}. Top 4 results are shown. N/A denotes relation facts that are not observed in KG.
}

\label{rankcase}
\end{table*}
}

{
\begin{table}[t]
\centering
\setlength\tabcolsep{2pt}
\begin{tabular}{c|cc|cc|cc}
Metrics & \multicolumn{2}{c}{CN15k} & \multicolumn{2}{|c}{NL27k} & \multicolumn{2}{|c}{PPI5k}\\
\hline
Dataset & F-1 & Accu. & F-1  & Accu. & F-1 & Accu.  \\
\hline
TransE & 23.4 & 67.9  & 65.1 &  53.4 & 83.2 & 98.5   \\
DistMult & 27.9 & 71.1   & 72.1 & 70.1 & 86.9 & 97.1 \\
ComplEx & 18.9 & 73.2 & 63.3 & 53.4 & 83.2 & 98.9  \\
URGE & 21.2 & 86.0 & 83.6 & 88.7 & 85.2 & 98.6 \\
\hline
\texttt{UKGE}$_{n-}$ & 23.6  & 86.1   & 64.4  & 65.5 & 92.7 & 99.3\\
\texttt{UKGE}$_{p-}$ &  26.2 & 88.7  & 89.7 & 93.4 & 94.2  & 99.3 \\
\hline
\texttt{UKGE$_{rect}$} & \textbf{28.8} & \textbf{90.4} & \textbf{92.3} & \textbf{95.2} & \textbf{95.1} & 99.4 \\
\texttt{UKGE$_{logi}$} & 25.9 & 90.1 & 88.4 & 93.0 & 94.5 & \textbf{99.5} \\
\end{tabular}
\caption{F-1 scores (\%) and accuracies (\%) of relation fact classification}
\label{classify}
\end{table}
}
\subsubsection{CN15k} 
CN15k is a subgraph of the commonsense KG ConceptNet.
This subgraph contains 15,000 entities and 241,158 uncertain relation facts in English.
The original scores in ConceptNet vary from 0.1 to 22, where 99.6\% are less than or equal to 3.0. 
For normalization, we first bound confidence scores to $x \in [0.1,3.0]$, and then applied the min-max normalization on $\log x$ to map them into [0.1, 1.0].

\subsubsection{NL27}
NL27k is extracted from NELL \cite{mitchell2018never}, an uncertain KG obtained from webpage reading. NL27k contains 27,221 entities, 404 relations, and 175,412 uncertain relation facts. In the process of min-max normalization, we search for the lower boundary from 0.1 to 0.9. We have found out that normalizing the confidence score to interval $[0.1,1]$ yields best results.

\subsubsection{PPI5k}
The Protein-Protein Interaction Knowledge Base STRING labels the interactions between proteins with the probabilities of occurrence. PPI5k is a subset of STRING that contains 271,666 uncertain relation facts for 4,999 proteins and 7 interactions. 

\par

In an uncertain KG, a relation fact is considered \emph{strong} if its confidence score $s_l$ is above a KG-specific threshold $\tau$.
Here we set $\tau=0.85$ for both CN15k and NL27k. We follow the instructions from \cite{szklarczyk2016string} and set $\tau=0.70$ for PPI5k. Under this setting, 20.4\% of relation facts in CN15k, 20.1\% of those in NL27k, and $12.4\%$ of those in PPI5k are considered strong. 
 
\subsection{Experimental Setup}

We split each dataset into three parts: 85\% for training, 7\% for validation, and 8\% for testing. 
To test if our model can correctly interpret negative links, we add the same amount of negative links as existing relation facts into the test sets.

We use Adam optimizer \cite{kingma2014adam} for training, for which we set the exponential decay rates $\beta_1=0.9$ and $\beta_2=0.99$. 
We report results for all models respectively based on their best hyperparameter settings.
For each model, the setting is
identified based on the validation set performance. We select among the following sets of hyper-parameter values:
learning rate $lr \in$ \{0.001, 0.005, 0.01\}, 
dimensionality $k \in$ \{64, 128, 256, 512\}, batch size $b \in$ \{128, 256, 512, 1024\},
The $L_2$ regularization coefficient $\lambda$ is fixed as 0.005. Training was stopped using early stopping based on MSE on the validation set, computed every 10 epochs. 
The best hyper-parameter combinations on CN15k and NL27k are $\{lr=0.001, k=128\}$ and $b=128$ for \texttt{UKGE$_{rect}$}, $b=512$ for \texttt{UKGE$_{logi}$}. On PPI5k they are $\{lr=0.001, k=128, b=256\}$ for both variants.

\subsection{Logical Rule Generation}
Our model requires additional input as logical rules for PSL reasoning. We heuristically create candidate logical rules by considering length-2 paths (i.e., ($E_1$,\texttt{$R_1$},${E_2}$) $\wedge$ ($E_2$,\texttt{$R_2$},${E_3}$) $\rightarrow$ ($E_1$,\texttt{$R_3$},${E_3}$)) and validate them by \emph{hit ratio,} i.e. the proportion of relation facts implied by the rule to be truly existent in the KG. The higher ratio implies that the rule is more convincing. When grounding out the logical rules, to guarantee the quality of the ground rules, we only adopt observed strong relation facts in our rule body. We eventually create 3 logical rules for CN15k, 4 for NL27k, and 1 for PPI5k. Table \ref{psl_rules} gives some examples of the logical rules and their hit ratios. 
How to systematically create more promising logical rules will be considered as future work. 

\subsection{Baselines}
Three types of baselines are considered in our comparison, which include (i) deterministic KG embedding models TransE \cite{bordes2013translating}, DistMult \cite{yang2014embedding} and ComplEx \cite{trouillon2016complex}, (ii) an uncertain graph embedding model URGE \cite{hu2017embedding}, and (iii) \texttt{UKGE}$_{n-}$~and \texttt{UKGE}$_{p-}$~that are two simplified versions of our model.
\begin{itemize}
\item Deterministic KG Embedding Models. TransE, DistMult, and ComplEx have demonstrated high performance on deterministic KGs.
Only the high-confidence relation facts from KGs are used for training.
For each KG, we have a KG-specific confidence score threshold $\tau$ to distinguish the high-confidence relation facts from the low-confidence ones, which will be discussed later in Section \ref{sec:classify}.
These models cannot predict confidence scores. We compare our methods to them only on the ranking and the classification tasks.
For the same reason, the early stopping is based on mean reciprocal rank (MRR) on the validation set.
We adopt the implementation given by \cite{trouillon2016complex} and choose the best hyper-parameters following the same grid search procedure. 
This implementation uses \cite{duchi2011adaptive} for optimization. The best hyper-parameter combinations on CN15k and NL27k are $b=1024$, \{$lr=0.01, k=128$\} for TransE and $\{lr=0.05, k=256\}$ for DistMult and ComplEx. On PPI5k they are $lr=0.1$, \{$k=128, b=512$\} for DistMult and \{$k=256, b=1024$\} for TransE and ComplEx.
\item Uncertain Graph Embedding Model. URGE is proposed very recently to embed uncertain graphs. However, it cannot deal with multiple types of relations in KGs, and it only produces node embeddings. We simply ignore relation types when applying URGE to our datasets. We adopt its first-order proximity version as our tasks focus on the edge relations between nodes.

\item Two Simplified Versions of Our Model. To justify the use of negative links and PSL reasoning in our model, we propose two simplified versions of \texttt{UKGE$_{rect}$}~, called \texttt{UKGE}$_{n-}$~and \texttt{UKGE}$_{p-}$.
In \texttt{UKGE}$_{n-}$, we only keep the observed relation facts and remove negative sampling, and in \texttt{UKGE}$_{p-}$, we remove PSL reasoning and use the MSE loss for unseen relation facts.
\end{itemize}

\subsection{Confidence Prediction} \label{prediction}
The objective of this task is to predict confidence scores of unseen relation facts. 

\subsubsection{Evaluation protocol} 
For each uncertain relation fact $(l,s_l)$ in the test set, we predict the confidence score of $l$ and report the mean squared error (MSE) and mean absolute error (MAE). 
\par

\subsubsection{Results}
Results are reported in Table \ref{mse}. Both our variants \texttt{UKGE$_{rect}$}~and \texttt{UKGE$_{logi}$}~outperform the baselines URGE, \texttt{UKGE}$_{n-}$, and \texttt{UKGE}$_{p-}$, since URGE only takes node proximity information and cannot model the rich relations between entities, and \texttt{UKGE}$_{n-}$~does not adopt negative sampling and cannot recognize negative links. The better results of \texttt{UKGE$_{rect}$}~than \texttt{UKGE}$_{p-}$~demonstrate that introducing PSL into embedding learning can enhance the model performance. Between the two model variants, \texttt{UKGE$_{rect}$}~results in smaller MSE and MAE than \texttt{UKGE$_{logi}$}. We notice that all the models achieve much smaller MSE on PPI5k than CN15k and NL27k. We hypothesize that this is because the much higher density of PPI5k facilitates embedding learning \cite{pujara2017sparsity}.

\subsection{Relation Fact Ranking}
The next task focuses on ranking tail entities in the right order for the query $(h, r, \underline{?t})$.

\subsubsection{Evaluation protocol} 
For a query $(h,r,\underline{?t})$, we rank all the entities in the vocabulary as tail candidates
and evaluate the ranking performance using the normalized Discounted Cumulative G ain (nDCG) \cite{liu2009learning}.
We define the gain in retrieving a relevant tail $t_0$ as the ground truth confidence score $s_{(h,r,t_0)}$.
We take the mean nDCG over the test query set as our ranking metric. We report the two versions of nDCG that use linear gain and exponential gain respectively. The exponential gain version puts stronger emphasis on highly relevant results.

\subsubsection{Results} 
Table \ref{rank2} shows the mean nDCG  over all test queries for all compared methods. 
Though TransE, DistMult, and ComplEx do not encode the confidence score information, they maximize the plausibility of all observed relation facts and therefore rank these existing relation facts high. 
We observe that DistMult and ComplEx have considerably better performance than TransE, as TransE does not handle \texttt{1-to-N} relations well. 
ComplEx embeds entities and relations in the complex domain and handles asymmetric relations better than DistMult. It achieves the best results among the deterministic KG embedding models on this task.
As \texttt{UKGE}$_{n-}$~removes negative sampling from the loss function, it cannot distinguish the negative links from existing relation facts and results in the worst performance. \texttt{UKGE}$_{p-}$~yields slightly worse performance than \texttt{UKGE$_{rect}$}. 
Besides ranking the existing relation facts highly, our models also preserve the order of the observed relation facts and thus achieve higher nDCG scores.
Both \texttt{UKGE$_{rect}$}~and \texttt{UKGE$_{logi}$}~outperform all the baselines under all settings, while \texttt{UKGE$_{logi}$}~yields higher nDCG on all three datasets than \texttt{UKGE$_{rect}$}. Considering the confidence prediction results of \texttt{UKGE$_{logi}$}~in Section \ref{prediction}, we hypothesize that the easy saturation of logistic function allows \texttt{UKGE$_{logi}$}~to better distinguish negative links from true relation facts, while this feature compromises its ability to fit confidence scores more precisely.

\subsubsection{Case study} Table \ref{rankcase} gives some examples of relation fact ranking results by \texttt{UKGE$_{logi}$}. 
Given a query $(h,r,\underline{?t})$, the top 4 predicted tails and true tails are given, sorted by their scores in descending order. The predictions are consistent with our common-sense. It is worth noting that some quite reasonable unseen relation facts such as \textit{hotel is used for relaxing}, can be predicted correctly. In other words, our proposed approach can be potentially used to infer new knowledge from the observed ones with reasonable confidence scores, which may shed light on  another line of future study.

\subsection{Relation Fact Classification} \label{sec:classify}

This last task is a binary classification task to decide whether a given relation fact $l$ is a \emph{strong} relation fact or not. A relation fact is considered \emph{strong} if its confidence score $s_l$ is above a KG-specific threshold $\tau$. The embedding models need to distinguish relation facts in the KG from negative links and high-confidence relation facts from low-confidence ones.

\subsubsection{Evaluation protocol} 
We follow a procedure that is similar to \cite{wang2014knowledge}. 
Our test set consists of relation facts from the KG and randomly sampled negative links equally. We divide the test cases into two groups, strong and weak/false, by their ground truth confidence scores. A test relation fact $l$ is strong when $l$ is in the KG and $s_l>\tau$, otherwise weak/false.
We fit a logistic regression classifier as a downstream classifier on the predicted confidence scores.

\subsubsection{Results} 
F-1 scores and accuracies are reported in Table \ref{classify}.
These results show that our two model variants consistently outperform all baseline models.
The deterministic KG models can distinguish the existing relation facts from negative links, but they do not leverage the confidence information and cannot recognize the high-confidence ones.
URGE does not encode the rich relations.
Although \texttt{UKGE}$_{n-}$~fits confidence scores in the KG, it cannot correctly interpret negative links as false. Consistent with the previous two tasks, the performance of \texttt{UKGE}$_{p-}$~is worse than \texttt{UKGE$_{rect}$}.



\section{Conclusion and Future Work}
To the best of our knowledge, this paper is the first work on embedding uncertain knowledge graphs. Our model \texttt{UKGE} effectively preserves both the relation facts and uncertainty information in the embedding space of KG. We propose two variants of our model and conduct extensive experiments on relation fact confidence score prediction, relation fact ranking, and relation fact classification. 
The results are very promising. For future work, we will study how to systematically generate reasonable logical rules and test its impact on embedding quality.
We are interested in extending \texttt{UKGE}~for uncertain knowledge extraction from text.

\subsection*{Acknowledgements}
This work is partially supported by NSF III-1705169, NSF CAREER Award 1741634, Snapchat gift funds, and PPDai gift fund.

\fontsize{9.0pt}{10.0pt} \selectfont
\bibliography{ref}
\bibliographystyle{aaai}
\end{document}